\documentclass[a4paper,conference]{IEEEtran}
\usepackage{cite}
\ifCLASSINFOpdf
   \usepackage[pdftex]{graphicx}
\else
\fi
\usepackage{amsmath}
  \usepackage[caption=false,font=normalsize,labelfont=sf,textfont=sf]{subfig}
\usepackage{url}
\begin{document}
%\title{Lightweight Human Pose Estimation Using Loss Weighted by Target Heatmap}

\title{Lightweight Human Pose Estimation Using Heatmap-Weighting Loss}
\author{\IEEEauthorblockN{Shiqi Li and Xiang Xiang}
\IEEEauthorblockA{MoE Key Lab of Image Information Processing and Intelligent Control\\
School of Artificial Intelligence and Automation,
Huazhong University of Science and Technology, China\\
$\mathsf{xex@hust.edu.cn}$}
}


%Telephone: (800) 555--1212, Fax: (888) 555--1212}


\maketitle

\begin{abstract}
Recent research on human pose estimation exploits complex structures to improve performance on benchmark datasets, ignoring the resource overhead and inference speed when the model is actually deployed. In this paper, we lighten the computation cost and parameters of the deconvolution head network in SimpleBaseline and introduce an attention mechanism that utilizes original, inter-level, and intra-level information to intensify the accuracy. Additionally, we propose a novel loss function called heatmap weighting loss, which generates weights for each pixel on the heatmap that makes the model more focused on keypoints. Experiments demonstrate our method achieves a balance between performance, resource volume, and inference speed. Specifically, our method can achieve 65.3 AP score on COCO test-dev, while the inference speed is 55 FPS and 18 FPS on the mobile GPU and CPU, respectively.
\end{abstract}


% \fi
\IEEEpeerreviewmaketitle

\section{Introduction}
Human pose estimation aims to locate the body joints from image or video data. It has received increasing attention from the computer vision community in the past couple of years and has been utilized in various applications such as animation\cite{animation}, sports\cite{sports,xiang2008intelligent}, and healthcare\cite{healthcare}. Due to the development of deep convolutional neural networks, recent works achieve significant progress on this topic\cite{hourglass, simplebaseline, cpn, hrnet}. 

In order to get higher accuracy, many existing methods usually comprise wide and deep network architectures because they could bring powerful representation capacity. This leads to tremendous parameters and a huge number of floating-point operations (FLOPs). However, the most common operating environments for pose estimation applications are resource-constrained edge devices such as smartphones or robots. The incremental computing cost and parameter memory are crucial barriers to the deployment of pose estimation. 

There is also some work focus on the lightweight human pose estimation method. Lite-HRNet\cite{litehrnet} follows High-Resolution Net (HRNet)\cite{hrnet} and reduces its parameters and computation cost. Lite-HRNet achieves extremely low parameters and FLOPs, however, the theoretical lightweight does not always mean fast in practice. The parallelism of the multi-branch structure does not reach the theoretical efficiency and the cross-resolution weight computation also slows down the inference of the model. It is even slower than the HRNet during actual running. Benefit from network architecture search (NAS) technology, ViPNAS\cite{vipnas} reaches a notable balance between accuracy and speed, but its performance while handling picture tasks on the CPU is not satisfactory. Other methods are short in terms of parameter quantity\cite{mspn}. Research in the lightweight human pose estimation area is still inadequate.

The SimpleBaseline\cite{simplebaseline} is an elegant and effective method for human pose estimation, they provide the capacity of deconvolution layers in this problem. Based on their successful work, we focus on the design of lightweight and simple network architecture for human pose estimation. First, we use a lightweight network MobileNetV3\cite{mobilenetv3} as the backbone in our model, rather than the huge  ResNet\cite{resnet}. As the latest model in MobileNet series\cite{mobilenetv1,mobilenetv2,mobilenetv3} MobileNetV3 has shown its performance in many computer vision tasks. Then we analyse the parameters and computation cost in the whole model and notice the deconvolution layers contain the dominant resources. Inspired by the design of the MobileNet series, we change the normal deconvolution layers to a depthwise deconvolution layer following a pointwise convolution layer. To compensate for the performance degradation caused by lightweighting, we introduce an attention mechanism. The attention block utilizes channel, spatial and original information of the input feature map to enhance the performance. Besides, we observe that Mean Square Error (MSE) loss is widely used by existing methods during the training stage. MSE loss just calculates the average error for different pixels on the heatmaps, without reflecting the inhomogenous from different locations. We propose a novel loss function called heatmap weighting loss to generate weights for every pixel on heatmaps that is used to calculate the final error. 

In short, our contribution can be summarized as followed.
\begin{itemize}
    \item We simplify the deconvolution head network in SimpleBaseline and present an attention mechanism that exploits channel, spatial and global representations to benefit the performance of the lightweight architecture.
    \item We propose a novel loss function - heatmap weighting loss that explores the information from ground truth heatmaps to improve keypoints localization.
\end{itemize}

\section{Related Work}
\subsection{Human Pose Estimation}
Human posture estimation has been an active research problem in the last decades. Before the rise of deep learning, human pose estimation was mainly based on graph structure models, using picture global features or handcraft filters to detect and localize the keypoints of human body\cite{traditional}. Since Toshev et al.\cite{deeppose} propose DeepPose, which uses the deep neural network to regress the keypoints locations, the neural network based approach becomes mainstream. Newell et al.\cite{hourglass} propose Hourglass, which consists of multi-stage stacked hourglass networks, it is a milestone on the MPII\cite{mpii} benchmark dataset. Chen et al.\cite{cpn} propose Cascaded Pyramid Network (CPN), which is the champion of COCO 2017 keypoint challenge. CPN uses a GlobalNet\cite{cpn} to estimate simple keypoints and handle complex keypoints with a RefineNet\cite{cpn}. Xiao et al.\cite{simplebaseline} propose SimpleBaseline, which is a simple but effective method composed of a backbone and several deconvolution layers. Sun et al.\cite{hrnet} present a novel High-Resolution Net to maintain high-resolution representation by connecting multi-resolution convolutions in parallel and conducting repeated multi-scale feature map fusion. Cai et al.\cite{prm} introduce Residual Steps Network (RSN) to learn more delicate local feature by intra-level information fusion, and propose Pose Refine Machine (PRM) to refine feature map by attention mechanism. This method is the winner of COCO 2019 keypoint challenge.

\subsection{Lightweight Network Design}
Design deep neural network architecture for achieving a trade-off between efficiency and accuracy has been an active research topic, especially in industrial research. In recent years, a large number of compact network architectures have been proposed. Xception\cite{xception} uses depthwise separable convolution operation to reduce parameters and computation. MobileNets series\cite{mobilenetv1, mobilenetv2, mobilenetv3} are also based on depthwise convolution and pointwise convolution. MobileNetV2\cite{mobilenetv2} present inverted residual block and MobileNetV3\cite{mobilenetv3} exploit NAS method searching a optimal structure that has better performance and fewer FLOPs. ShuffleNet\cite{shufflenet} proposes a channel shuffle operation for information interaction between feature map groups. ShuffleNetV2\cite{shufflenetv2} studies the lightweight network design principles and modifies the ShuffleNet according to the proposed guidelines. GhostNet\cite{ghostnet} investigates the correlation and redundancy between feature maps and presents a cheap operation to reduce the computation of pointwise convolution. The development of lightweight networks has great significance for our work, on the one hand, we can directly use a more powerful lightweight network as a backbone in our model, on the other hand, we can draw on the idea of compact network design in our downstream tasks.

\subsection{Attention Mechanism}
Attention mechanism has shown powerful performance in many computer vision tasks such as image classification, scene segmentation, objection detection, and optical character recognition (OCR). Wang et al.\cite{nonlocal} propose Non-Local Networks which introduce self-attention to capture long-range dependencies. Hu et al.\cite{se} present Squeeze-and-Excitation Networks (SENet), using weight learned from global information to excite or suppress different channels, which is the champion of ILSVRC 2017. Li et al.\cite{sknet} expand SENet and propose Selective Kernel Networks (SKNet), using feature maps generated by multiple kernel sizes to gather attention value. Woo et al.\cite{cbam} propose Convolutional Block Attention Module (CBAM) to exploit both channel and spatial information to refine the feature map. Most attention modules can be easily embedded in our pose estimation network without introducing too much extra computation while improving the performance of the network.

\section{Method}
\subsection{Lightweight Deconvolution Head Network}
Deconvolution head network is a different method to rehabilitate high-resolution feature map which is introduced in SimpleBaseline. Before that, upsampling is the dominant technology to recover high-resolution feature map\cite{hourglass, cpn}, and some convolution layers added to adjust the generated feature map. In the deconvolution head network, a deconvolution layer that could learn the upsampling parameters from input data during training combines upsampling and convolution parameters in a single operation. 

The standard deconvolution layer is parameterized by the convolution kernel $K$ of size $D_K\times D_K\times C_{in}\times C_{out}$, where $D_K$ is the spatial dimension of the kernel assumed to be square and $C_{in}$ is the number of input channels, $C_{out}$ is the number of output channels, and $l$ is the number of deconvolution layers. Despite the first layer for channel compression and the last layer for generating output feature map, the total number of parameters in a deconvolution head network is:
\begin{equation}
    \sum_i^l D_{K_i}\times D_{K_i}\times C_{in_i}\times C_{out_i}
\end{equation}

The computational cost of a deconvolution layer multiplicatively depends on the size of kernel $D_K \times D_K$, the number of input channels $C_{in}$, the number of output channels $C_{out}$, and the size of feature maps $W\times H$. Therefore, the total computation in a $l$ layer deconvolution head network is:
\begin{equation}
    \sum_i^l D_{K_i}\times D_{K_i}\times C_{in_i}\times C_{out_i} \times W_i \times H_i
\end{equation}
\begin{figure}[!t]
    \centering
    \includegraphics[width=0.9\linewidth]{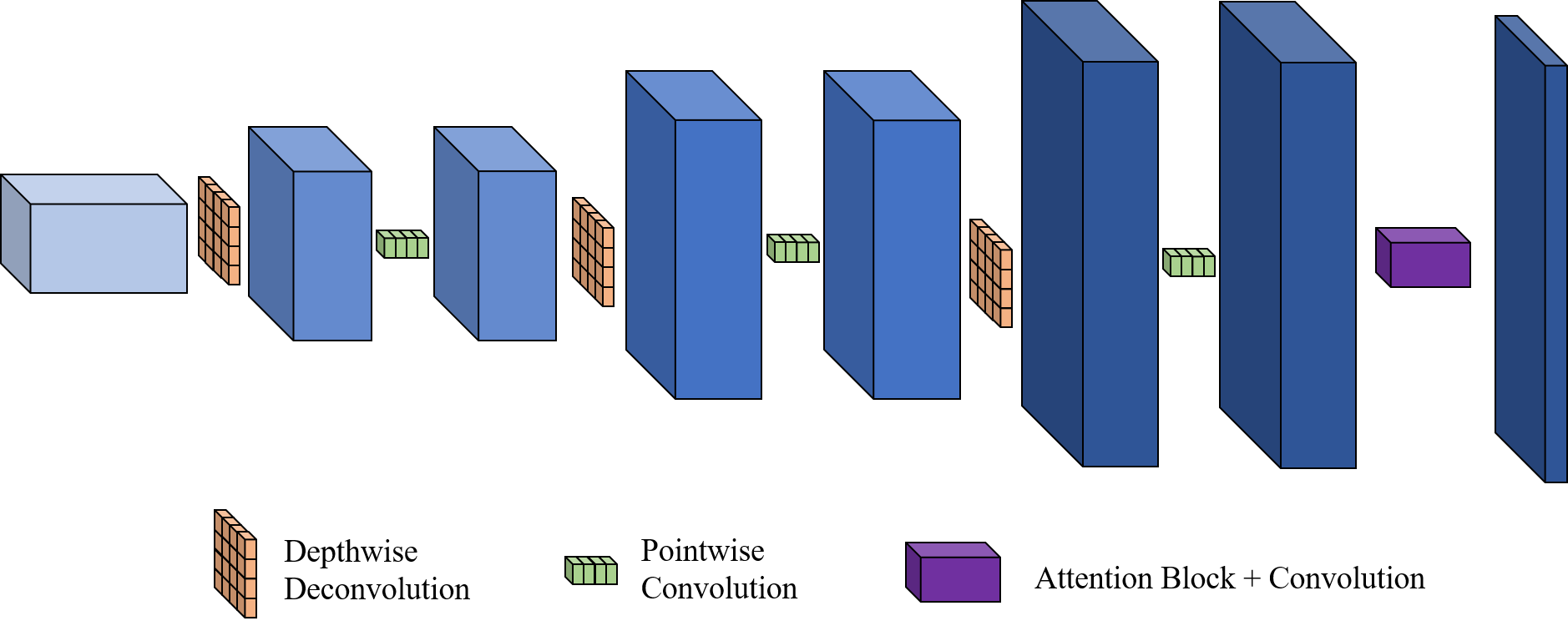}
    \caption{Illustration of lightweight deconvolution head network.}
    \label{network}
\end{figure}

Based on the design in MobileNets, we replace the standard deconvolution layer with a depthwise deconvolution layer following a pointwise convolution layer. Fig. \ref{network} illustrates the structure of our lightweight deconvolution head network. Depthwise deconvolution applies a single filter on each channel for upsampling and producing the new representations in high-resolution feature map, while pointwise convolution provides a bridge for information fusion across channels. The influence of the output channels on the parameters and computation is eliminated in depthwise deconvolution, and in pointwise convolution, the kernel size will no longer be a factor. The computation cost of our new deconvolution head network can be written as:
\begin{equation}
    \sum_i^l D_{K_i}\times D_{K_i}\times C_{in_i}\times W_i \times H_i + C_{in_i}\times C_{out_i} \times W_i \times H_i
\end{equation}
The number of parameters is proportional to the computation cost, just remove the size of feature map.
Following the setting in SimpleBaseline, we set the layers to 3, kernel size to 4, and the number of channels to 256 in all layers. Thus we can get a reduction in computation cost of:
\begin{equation}
\begin{aligned}
    &\frac{\sum_i^3 4\times 4\times 256\times W_i \times H_i + 256\times 256 \times W_i \times H_i}{\sum_i^3 4\times 4\times 256\times 256 \times W_i \times H_i} \\
    =&\frac{17}{256}\approx\frac{1}{16}.
\end{aligned}
\end{equation}

\subsection{Attention Mechanism}
To improve the capacity of our lightweight deconvolution head network, we introduce an attention mechanism. Motivated by the work of SENet, we use an adaptive average pooling to aggregate spatial information in each channel. After generating channel representations, two pointwise convolution layers are used to calculate the correlation between channels. To reduce the parameters, the input feature dimension $C$ is squeezed to $C/r$ after the first convolution, where $r$ is the squeeze ratio. Before finally broadcasting along spatial dimension and multiplying with the original feature map, a hard sigmoid function is utilized to normalize the weight. Define the feature map as $f$, a convolution layer with $p$ output channels and kernel size of $m\times n$ as $\text{F}^{m\times n}_{p}$, the hard sigmoid as $\sigma(\cdot)$, the global average pooling as $\text{GAP}(\cdot)$, and the element-wise multiplication as $\odot$. The channel attention $\mathcal{H}_{channel}(\cdot)$ is computed as:
\begin{equation}
    \mathcal{H}_{channel}(f) = f \odot \sigma(\text{F}^{1\times1}_{C}(\text{F}^{1\times1}_{C/r}(\text{GAP}(f))))
\end{equation}

\begin{figure}[!t]
    \centering
    \includegraphics[width=0.95\linewidth]{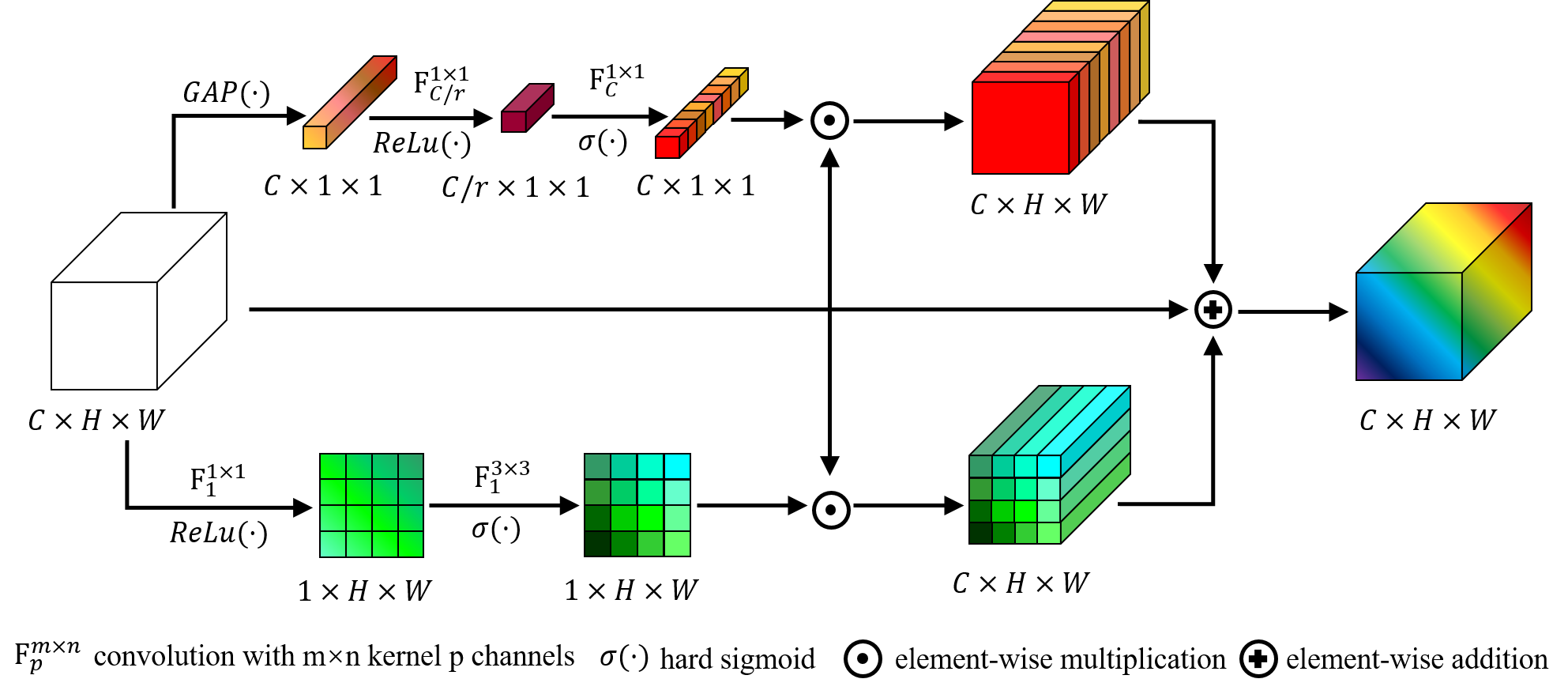}
    \caption{Illustration of attention architecture.}
    \label{attention}
\end{figure}

Since human pose estimation is a space-sensitive task, a spatial attention mechanism can help to better mine the information in spatial dimension. To converge information in different channels, we apply a pointwise convolution to get a linearly combined representation. After that, a standard convolution is used to produce the spatial attention map. A hard sigmoid layer is also used to rescale the activation values. Following the definition of the channel portion, the spatial attention $\mathcal{H}_{spatial}(\cdot)$ is written as: 
\begin{equation}
    \mathcal{H}_{spatial}(f) = f \odot \sigma(\text{F}^{3\times3}_{1}(\text{F}^{1\times1}_{1}(f))) 
\end{equation}

For the arrangement of attention modules, we use a parallel structure, as shown in Fig. \ref{attention}. Channel attention and spatial attention are applied to the feature map individually and an identity connection is also employed. The identity mapping protects local features from being corrupted by dimensional compression in the attention block. Meanwhile, channel attention exploits inter-level representation to benefit keypoints identification and spatial attention improves keypoints by optimizing spatial information. Finally, the feature maps in three branches are added to get the output feature. Our attention mechanism can be formulated as:
\begin{equation}
    f_{out}=f_{in} + \mathcal{H}_{channel}(f_{in}) + \mathcal{H}_{spatial}(f_{in}) \\
\end{equation}

\subsection{Heatmap Weighting Loss}
\begin{figure}[!t]
    \centering
    \includegraphics[width=\linewidth]{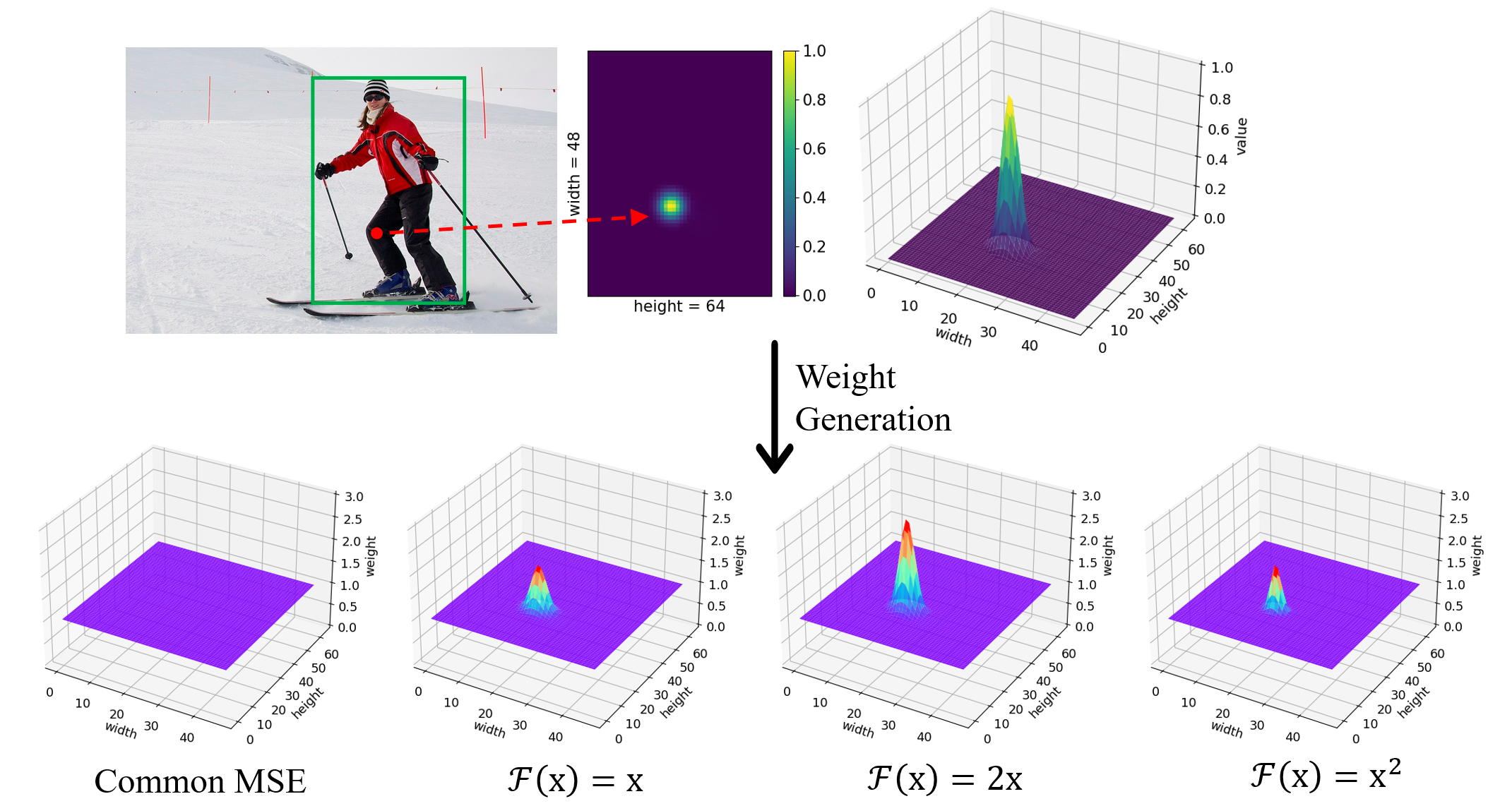}
    \caption{Proposed heatmap weighting loss.}
    \label{loss}
\end{figure}

In the training phase, MSE is commonly used as the loss function\cite{hourglass, simplebaseline, cpn, hrnet}. Beyond the previous work, we argue that the weight of the pixels at different locations on the heatmaps should be various. The loss function should focus more on the points with larger values on the heatmaps, which means devoting more attention to those pixels that are closer to the keypoints. Based on this thinking, we propose heatmap weighting loss which generates weight for each pixel by adopting the transformation function on the ground truth heatmap, as shown in Fig. \ref{loss}. Our loss function presents as:
\begin{equation}
    L=\frac{1}{J}\sum^J_{j=1}\left[\mathcal{F}(P^{GT}_j) + 1\right]\odot\Vert P_j-P^{GT}_j \Vert_2
\end{equation}
where $J$ is the number of keypoints, $P$ and $P^{GT}$ are predicted heatmap and ground truth heatmap respectively, $\mathcal{F}(\cdot)$ is weight generation function, and $\odot$ denotes element-wise multiplication.

For the selection of weight generation function, in consideration of the target heatmaps are usually sampled from a two-dimensional Gaussian distribution\cite{heatmap}. We select several convex functions on $[0,1]$ interval such as linear function, power function, and exponential function to generate the weights and evaluate their performance. See Section \ref{subsec:loss} for details.

\section{Experiments}
We evaluate our method on COCO\cite{coco} and MPII\cite{mpii} datasets. Following the two-stage top-down paradigm\cite{simplebaseline, hrnet}, our approach estimates keypoints confidence maps from the detected person boundary box. We report the comparisons with other methods and also conduct inference speed test and ablation studies on these datasets. The project page \footnote{\url{https://github.com/HAIV-Lab/ICPR22w}} has been set up with programs available.

\subsection{Experimental Setup}
\subsubsection{Dataset and metric}
COCO dataset contains over 200K images and 250K person instances with 17 keypoints. Our model is trained on the train2017 set (includes 57K images and 150K person instances) and validated on the val2017 set (includes 5K images). MPII includes around 25K images containing over 40K people with annotated body joints. We adopt object keypoint similarity (OKS) based mean average precision (AP) to evaluate the performance on COCO, for MPII, the standard PCKh (head-normalized percentage of correct keypoints) is used.  

\subsubsection{Image cropping}
The human box is expanded in height or width to reach a fixed aspect ratio, e.g., height:width=4:3. Then we crop the box area from the image and finally resize it to a certain resolution such as $256\times192$ for COCO and $256\times256$ for MPII.

\subsubsection{Data augmentation}
We apply a series of data augmentation operations to make the model learn scale invariance and rotation invariance. Augmentation strategy includes flip, random rotation ($[-80^\circ, 80^\circ]$), random scale ([0.5, 1.5]), and half body augmentation which keeps only the upper or the lower body at random for COCO.

\subsubsection{Training}
We use Adam optimizer mini-batch size 64 to train our network, the initial learning rate is 5e-4 and reduced by a factor of 10 at the 170th and 200th epoch. The training process is terminated within 210 epochs. The warm-up strategy with a liner warm ratio of 0.001 is used in  the beginning 500 iterators\cite{mmpose2020}. Our experiments are conducted on a single NVIDIA 3090 GPU. 

\subsubsection{Testing}
In order to minimize the variance of prediction, following the normal practice\cite{hourglass}, we average the heatmaps predicted from the original and flipped images. A quarter of a pixel offset in the direction from the highest response to the second-highest response is applied before the predicted keypoint locations transform back to the original coordinate space of the input image.

\subsection{Results}

\subsubsection{COCO val}
\begin{figure*}[!t]
    \centering
    \includegraphics[width=0.95\linewidth]{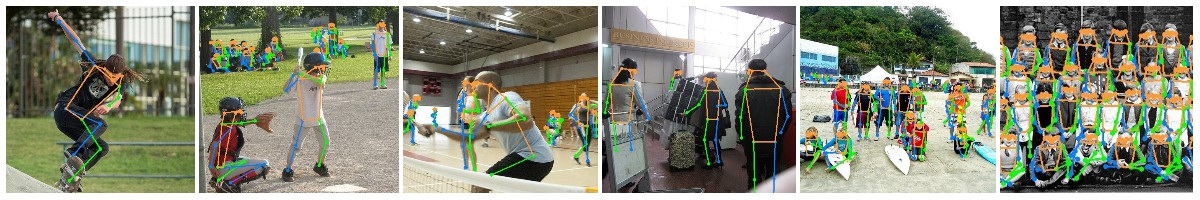}
    \caption{Prediction results on COCO val.}
    \label{vis}
\end{figure*}

\begin{table*}[!t]
%% increase table row spacing, adjust to taste
\renewcommand{\arraystretch}{1.5}
\caption{Comparison on COCO val2017 set.}
\label{cocoval}
\centering
\begin{tabular}{llccccccccc}
\hline
\multicolumn{1}{c}{Method} & \multicolumn{1}{c}{Backbone} & Input Size & \#Params & GFLOPs & AP & AP$^{50}$ & AP$^{75}$ & AP$^M$ & AP$^L$ & AR \\
\hline
SimpleBaseline\cite{simplebaseline} & MobileNetV2 & $256\times192$ & 9.6M & 1.59 & 64.6 & 87.4 & 72.3 & 61.1 & 71.2 & 70.7 \\
SimpleBaseline & MobileNetV2 & $384\times288$ & 9.6M & 3.57 & 67.3 & 87.9 & 74.3 & 62.8 & 74.7 & 72.9 \\
SimpleBaseline & MobileNetV3 & $256\times192$ & 8.7M & 1.47 & 65.9 & 87.8 & 74.1 & 62.6 & 72.2 & 72.1 \\
SimpleBaseline & ShuffleNetV2 & $256\times192$ & 7.6M & 1.37 & 59.9 & 85.4 & 66.3 & 56.5 & 66.2 & 66.4 \\
SimpleBaseline & ShuffleNetV2 & $384\times288$ & 7.6M & 3.08 & 63.6 & 86.5 & 70.5 & 59.5 & 70.7 & 69.7 \\
ViPNAS\cite{vipnas} & MobileNetV3 & $256\times192$ & 2.8M & 0.69 & 67.8 & 87.2 & 76.0 & 64.7 & 74.0 & 75.2 \\
Small HRNet\cite{litehrnet} & HRNet-W16 & $256\times192$ & 1.3M & 0.54 & 55.2 & 83.7 & 62.4 & 52.3 & 61.0 & 62.1 \\
Small HRNet & HRNet-W16 & $384\times288$ & 1.3M & 1.21 & 56.0 & 83.8 & 63.0 & 52.4 & 62.6 & 62.6 \\
Lite-HRNet\cite{litehrnet} & Lite-HRNet-18 & $256\times192$ & 1.1M & 0.20 & 64.8 & 86.7 & 73.0 & 62.1 & 70.5 & 71.2 \\
Lite-HRNet & Lite-HRNet-18 & $384\times288$ & 1.1M & 0.45 & 67.6 & 87.8 & 75.0 & 64.5 & 73.7 & 73.7 \\
Lite-HRNet & Lite-HRNet-30 & $256\times192$ & 1.8M & 0.31 & 67.2 & 88.0 & 75.0 & 64.3 & 73.1 & 73.3 \\
Lite-HRNet & Lite-HRNet-30 & $384\times288$ & 1.8M & 0.70 & 70.4 & 88.7 & 77.7 & 67.5 & 76.3 & 76.2 \\
\hline
Ours & MobileNetV3 & $256\times192$ & 3.1M & 0.58 & 65.8 & 87.7 & 74.1 & 62.6 & 72.4 & 72.1 \\
Ours & MobileNetV3 & $384\times288$ & 3.1M & 1.30 & 69.9 & 88.8 & 77.5 & 66.0 & 76.7 & 75.5 \\
\hline
\end{tabular}
\end{table*}
Table \ref{cocoval} reports the comparison results of our method and other state-of-the-art methods. Our method, trained from input resolution with $256\times192$, achieves a 65.8 AP score. Compared to ShuffleNetV2, our method improves AP by 5.9 points with $42\%$ FLOPs. We use about $35\%$ of the parameters and computational cost while obtaining a gain of 1.2 AP points compared to MobileNetV2. Our method matches the performance of SimpleBaseline using MobileNetV3 as the backbone, with only $35\%$ FLOPs and $39\%$ parameters. Compared to Small HRNet-W16\cite{litehrnet}, we improve 10.6 AP score. Our method also outperforms the Lite-HRNet-18 with 1.0 AP points. There are little gaps (1.3 points and 2.0 points) between our approach and the state-of-the-art methods, Lite-HRNet-30 and ViPNAS, respectively, but our method achieves a much higher inference speed. See Section \ref{subsec:speed} for details. It is worth noting that the goal of this work is to introduce a simple lightweight human pose estimation network with high inference speed rather than improving the score of existing SOTA approache on benchmark datasets. For the input size of $384\times288$, our method achieves 69.9 AP score. Some prediction results are visualized in Fig. \ref{vis}.

\subsubsection{COCO test-dev}
Table \ref{cocotest} reports the comparison results of our models and other lightweight methods. Our networks achieve 65.3 and 69.2 with the input size of $256\times192$ and $384\times288$, respectively. Our approach outperforms other methods except for Lite-HRNet-30 in AP scores, and our method is closer to SOTA with larger input size.
\begin{table*}[!t]
%% increase table row spacing, adjust to taste
\renewcommand{\arraystretch}{1.5}
\caption{Comparison on COCO test-dev2017 set.}
\label{cocotest}
\centering
\begin{tabular}{llccccccccc}
\hline
\multicolumn{1}{c}{Method} & \multicolumn{1}{c}{Backbone} & Input Size & \#Params & GFLOPs & AP & AP$^{50}$ & AP$^{75}$ & AP$^M$ & AP$^L$ & AR \\
\hline
SimpleBaseline\cite{simplebaseline} & MobileNetV2 & $256\times192$ & 9.6M & 1.59 & 64.1 & 89.4 & 71.8 & 60.8 & 69.8 & 70.1 \\
SimpleBaseline & MobileNetV2 & $384\times288$ & 9.6M & 3.57 & 66.8 & 90.0 & 74.0 & 62.6 & 73.3 & 72.3 \\
SimpleBaseline & ShuffleNetV2 & $256\times192$ & 7.6M & 1.37 & 59.5 & 87.4 & 66.0 & 56.6 & 64.7 & 66.0 \\
SimpleBaseline & ShuffleNetV2 & $384\times288$ & 7.6M & 3.08 & 62.9 & 88.5 & 69.4 & 58.9 & 69.3 & 68.9 \\
Small HRNet\cite{litehrnet} & HRNet-W16 & $384\times288$ & 1.3M & 1.21 & 55.2 & 85.8 & 61.4 & 51.7 & 61.2 & 61.5 \\
Lite-HRNet\cite{litehrnet} & Lite-HRNet-18 & $256\times192$ & 1.1M & 0.20 & 63.7 & 88.6 & 71.1 & 61.1 & 68.6 & 69.7 \\
Lite-HRNet & Lite-HRNet-18 & $384\times288$ & 1.1M & 0.45 & 66.9 & 89.4 & 74.4 & 64.0 & 72.2 & 72.6 \\
Lite-HRNet & Lite-HRNet-30 & $256\times192$ & 1.8M & 0.31 & 66.7 & 88.9 & 74.9 & 63.9 & 71.9 & 72.7 \\
Lite-HRNet & Lite-HRNet-30 & $384\times288$ & 1.8M & 0.70 & 69.7 & 90.7 & 77.5 & 66.9 & 75.0 & 75.4 \\
\hline
Ours & MobileNetV3 & $256\times192$ & 3.1M & 0.58 & 65.3 & 89.7 & 73.4 & 62.6 & 70.4 & 71.3 \\
Ours & MobileNetV3 & $384\times288$ & 3.1M & 1.30 & 69.2 & 90.6 & 76.9 & 65.8 & 74.9 & 74.7 \\
\hline
\end{tabular}
\end{table*}

\subsubsection{MPII val}
Comprehensive results on MPII are reported in Table 3. Our method achieves 85.9 PCKh with a standard $256\times256$ input size, outperforms the ShuffleNetV2, MobileNetV2 by 3.1 and 0.5 points, respectively.
\begin{table*}[!t]
%% increase table row spacing, adjust to taste
\renewcommand{\arraystretch}{1.5}
\caption{Comparison on MPII val set.}
\label{mpii}
\centering
\begin{tabular}{llcccccccccc}
\hline
\multicolumn{1}{c}{Method} & \multicolumn{1}{c}{Backbone} & \#Params & GFLOPs & Head & Shoulder & Elbow & Wrist & Hip & Knee & Ankle & PCKh \\
\hline
SimpleBaseline\cite{simplebaseline} & MobileNetV2 & 9.6M & 2.12 & 95.3 & 93.5 & 85.8 & 78.5 & 85.9 & 79.3 & 74.4 & 85.4 \\
SimpleBaseline & ShuffleNetV2 & 7.6M & 1.83  & 94.6 & 92.4 & 83.0 & 75.6 & 82.8 & 75.9 & 69.2 & 82.8 \\
Lite-HRNet\cite{litehrnet} & Lite-HRNet-18 & 1.1M & 0.27 & 96.1 & 93.7 & 85.5 & 79.2 & 87.0 & 80.0 & 75.1 & 85.9 \\
Lite-HRNet & Lite-HRNet-30 & 1.8M & 0.42 & 96.3 & 94.7 & 87.0 & 80.6 & 87.1 & 82.0 & 77.0 & 87.0 \\
\hline
Ours & MobileNetV3 & 3.1M & 0.77 & 95.6 & 93.9 & 85.1 & 79.5 & 86.3 & 80.4 & 75.5 & 85.9 \\
\hline
\end{tabular}
\end{table*}

\subsection{Weight Generation Function}
\label{subsec:loss}
We experimentally study the influence of heatmap weighting loss. We evaluate different weight generation functions and compare them with the traditional MSE loss. Table \ref{lossablation} reports the results on COCO val with the input size of $256\times192$.
\begin{table}[!t]
%% increase table row spacing, adjust to taste
\renewcommand{\arraystretch}{1.5}
\caption{Comparison of different weight generation functions.}
\label{lossablation}
\centering
\begin{tabular}{ccccc}
\hline
Weight Generation Function & AP & AP$^{50}$ & AP$^{75}$ & AR \\ 
\hline
None & 65.56 & 87.36 & 73.97 & 71.65 \\ 
$\mathcal{F}(x)=x$ & 65.83 & 87.70 & 74.06 & 72.06 \\
$\mathcal{F}(x)=2x$ & 65.59 & 87.37 & 74.01 & 71.90 \\
$\mathcal{F}(x)=x^2$ & 65.65 & 87.70 & 73.96 & 71.81 \\
$\mathcal{F}(x)=e^x$ & 65.70 & 87.66 & 73.74 & 71.79 \\
\hline
\end{tabular}
\end{table}

We can find that the AP score got the most significant gain of 0.27 points when using the most simple endomorphism as the weight generation function. When we increase the transformation ratio in linear mapping, the gain becomes negligible (only 0.03 AP score). We also attempt the square and exponential functions that obtain gains of 0.09 and 0.14 points, respectively. Although the gain from our heatmap weighting loss function is not significant now, it provides a new optimization idea for human pose estimation. The notion of our heatmap weighting loss can be considered as a kind of adjustment of the handcrafted two-dimensional Gaussian distribution heatmap to make it approximate the actual distribution of the human keypoint.

\subsection{Inference Speed}
\label{subsec:speed}
Operations (OPs) denote the number of computations required for the forward inference of the model and reflect the demand for hardware computational units. While float is the most commonly used data type, FLOPs becomes the general reference for evaluating model size. The actual inference speed on hardware is a complex issue. It is not only affected by the amount of computation, but also by many factors such as the amount of memory access, hardware characteristics, and system environment. For this reason, we study the actual inference speed of different human pose estimation models on an edge device.

\begin{figure}[!t]
    \centering
    \includegraphics[width=\linewidth]{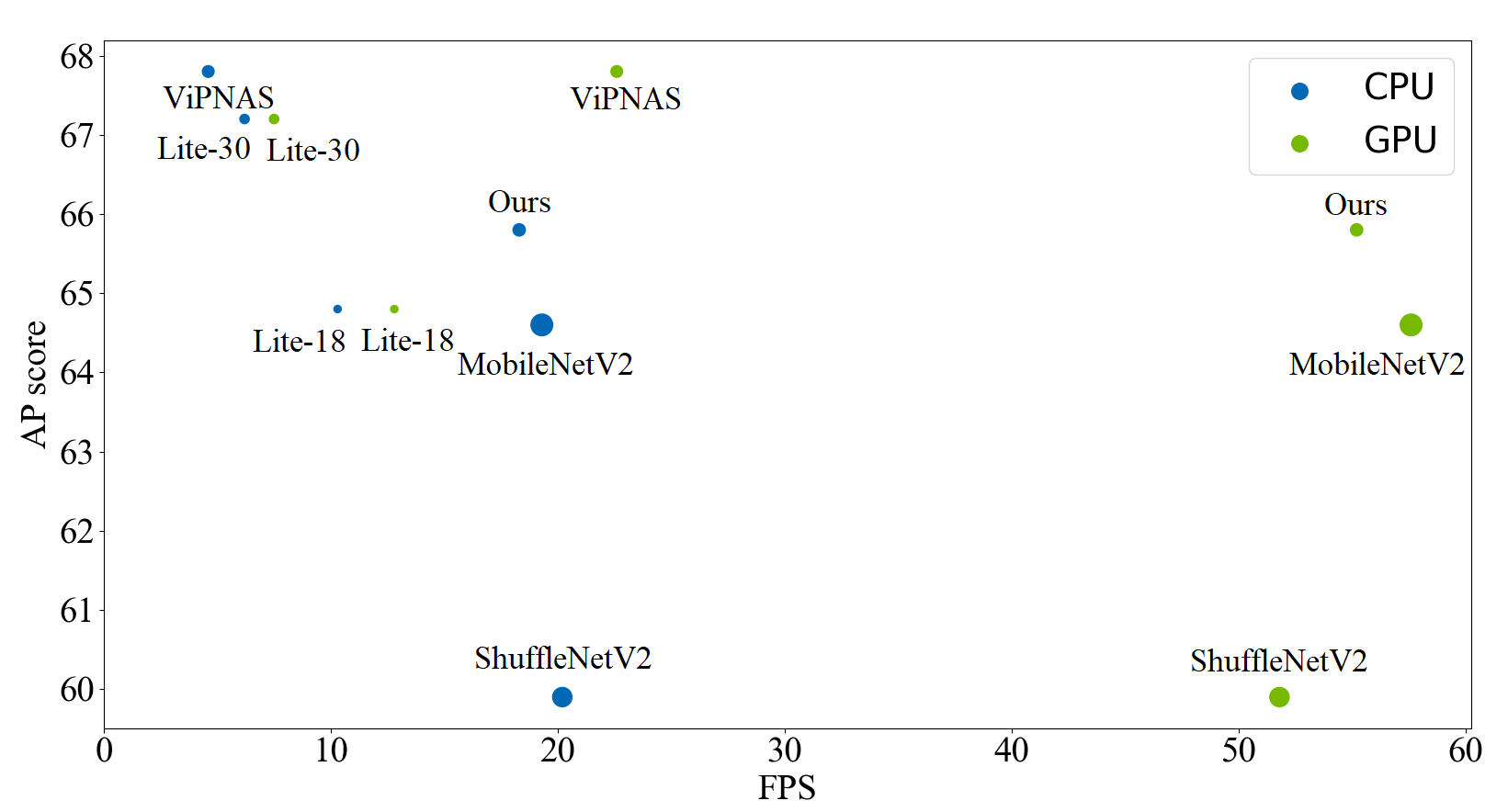}
    \caption{Speed-accuracy trade-off on COCO val. The circular area represents the number of parameters of the corresponding model.}
    \label{speed-pic}
\end{figure}

Our test platform is a laptop equipped with Intel Core i7-10750H CPU and NVIDIA GTX 1650Ti (Notebooks) GPU. We perform 50 rounds of inference with each model and calculate the average iterations per second. To reduce the error, the initial 5 iterations are not included in the statistics. The time for data pre-processing is eliminated and we keep the same input feature map from COCO val for each model with $256\times192$ resolution. We conduct several sets of experiments and calculate the average FPS as the final result. Table \ref{speed} and Fig. \ref{speed-pic} summarize the results of our speed experiment.

Our experiments confirm that FLOPs do not fully respond to inference speed. The GFLOPs of Lite-HRNet-18 and Lite-HRNet-30 are impressive 0.20 and 0.31, respectively. However, dragged down by the multi-branch parallel convolution architecture, the lite-HRNet series only achieves about 10 FPS on both CPU and GPU platforms. Their inference speed on GPU device is even slower than the speed of MobileNetV2, ShuffleNetV2, and our method on CPU platform. ViPNAS, the most accurate method, reaches 22.6 FPS on the GPU, but less than 5 FPS on the CPU. Our method achieves 55.3 FPS on GPU, which is 7.4 times higher than Lite-HRNet-30 and 2.4 times higher than ViPNAS. On the CPU platform, the FPS of our method is 18.3, about 4 times higher than that of ViPNAS. Our approach achieves higher accuracy while being essentially the same speed as MobileNetV2 and ShuffleNetV2. Inference speed experiments demonstrate that our method is more friendly to the mobile device and better suited to real-world applications.

\begin{table}[!t]
%% increase table row spacing, adjust to taste
\renewcommand{\arraystretch}{1.5}
\caption{comprarison of inference speed.}
\label{speed}
\centering
\begin{tabular}{lcccc}
\hline
\multicolumn{1}{c}{Model} & AP & FPS(GPU) & FPS(CPU) & GFLOPs \\ \hline
MobileNetV2\cite{mobilenetv2} & 64.6 & 57.6 & 19.3 & 1.59\\ 
ShuffleNetV2\cite{shufflenetv2} & 59.9 & 51.8 & 20.2 & 1.37\\ 
ViPNAS\cite{vipnas} & 67.8 & 22.6 & 4.6 & 0.69\\ 
Lite-HRNet18\cite{litehrnet} & 64.8 & 12.8 & 10.3 & 0.20\\ 
Lite-HRNet30\cite{litehrnet} & 67.2 & 7.5 & 6.2 & 0.31\\ \hline
Ours & 65.8 & 55.2 & 18.3 & 0.58\\
\hline
\end{tabular}
\end{table}

\subsection{Ablation Study}
\label{subsec:ablation}
In this part, we validate the improvement of our attention architecture by performing ablation experiments on COCO val set with $256\times192$ input size, as shown in Table 7.   We compare the prior attention mechanism SE-Block, CBAM, and our attention module. 

\begin{table}[!t]
%% increase table row spacing, adjust to taste
\renewcommand{\arraystretch}{1.5}
\caption{Comparison of different attention blocks.}
\label{attentionablation}
\centering
\begin{tabular}{ccccc}
\hline
Attention & AP & AP$^{50}$ & AP$^{75}$ & AR \\ 
\hline
None & 65.29 & 87.44 & 73.28 & 71.53 \\ 
SE-Block\cite{se} & 65.71 & 87.57 & 74.10 & 71.93 \\
CBAM\cite{cbam} & 65.77 & 87.92 & 74.19 & 71.02 \\
\hline
Ours & 65.83 & 87.70 & 74.06 & 72.06 \\
\hline
\end{tabular}
\end{table}
Our approach outperforms the other two modules in AP score, which implies trimming the common attention block designed for image classification before applying it to a certain task is necessary.

\section{Conclusion}
In this paper, we propose a lightweight human pose estimation network with high inference speed by simplifying the deconvolution head network in SimpleBaseline and applying an attention mechanism that exploits channel, spatial and global representations to improve the performance. We also present a novel loss function, heatmap weighting loss, which explores the information from ground truth heatmaps to benefit keypoints localization. Our method achieves a balance between performance, resource volume, and inference speed. We hope that this work gives a new idea for lightweight human pose estimation.


%\section*{Acknowledgment}




%\bibliographystyle{IEEEtran}
%\begin{thebibliography}{1}
%\end{thebibliography}
\newpage
\bibliographystyle{IEEEtran}
\bibliography{IEEEabrv,ref}

\end{document}